\documentclass[runningheads]{llncs}
\usepackage{graphicx}
\usepackage{comment}
\usepackage{amsmath,amssymb} 
\usepackage{xcolor}
\usepackage{xspace}
\usepackage{multirow}

\usepackage{wrapfig}
\usepackage[printwatermark]{xwatermark}
\newwatermark*[page=1,color=gray!60,fontfamily=cmr,fontseries=m,scale=0.4,xpos=0,ypos=120]{This paper has been accepted for publication at the \\ European Conference on Computer Vision (ECCV), Online, 2020}

\newcommand{\eg}{\emph{e.g.}\xspace}

\begin{document}
\pagestyle{headings}
\mainmatter

\title{Learning to Exploit Multiple Vision Modalities by Using Grafted Networks} 

\titlerunning{Learning to Exploit Multiple Vision Modalities by Using Grafted Networks}

\author{Yuhuang Hu\orcidID{0000-0002-5543-7619} \and
Tobi Delbruck\orcidID{0000-0001-5479-1141} \and
Shih-Chii Liu\orcidID{0000-0002-7557-045X}}
\authorrunning{Y. Hu et al.}
%
\institute{Institute of Neuroinformatics, University of Z\"urich and ETH Z\"urich, Switzerland \email{\{yuhuang.hu, tobi, shih\}@ini.uzh.ch}}

\maketitle

\begin{abstract}
Novel vision sensors such as thermal, hyperspectral, polarization, and event cameras provide information that is not available from conventional intensity cameras. An obstacle to using these sensors with current powerful deep neural networks is the lack of large labeled training datasets. 
This paper proposes a Network Grafting Algorithm (NGA),
where a new front end network driven by unconventional visual inputs replaces the front end network of a pretrained deep network that processes intensity frames. 
The self-supervised training uses only synchronously-recorded intensity frames and novel sensor data to maximize feature similarity between the pretrained network and the grafted network.
We show that the enhanced grafted network 
reaches competitive average precision (AP$_{50}$) scores to the pretrained network on an object detection task using thermal and event camera datasets, with no increase in inference costs. Particularly, the grafted network driven by thermal frames showed a relative improvement of 49.11\% over the use of intensity frames. 
The grafted front end has only 5--8\% of the total parameters and can be trained in a few hours on a single GPU equivalent to 5\% of the time that would be needed to train the entire object detector from labeled data.
NGA allows new vision sensors to capitalize on previously pretrained powerful deep models, saving on training cost and widening a range of applications for novel sensors.
\end{abstract}

\section{Introduction}\label{sec:intro}

\begin{figure}[ht]
\centering
\begin{minipage}[t]{0.43\linewidth}
    \begin{center}
    \includegraphics[width=\linewidth]{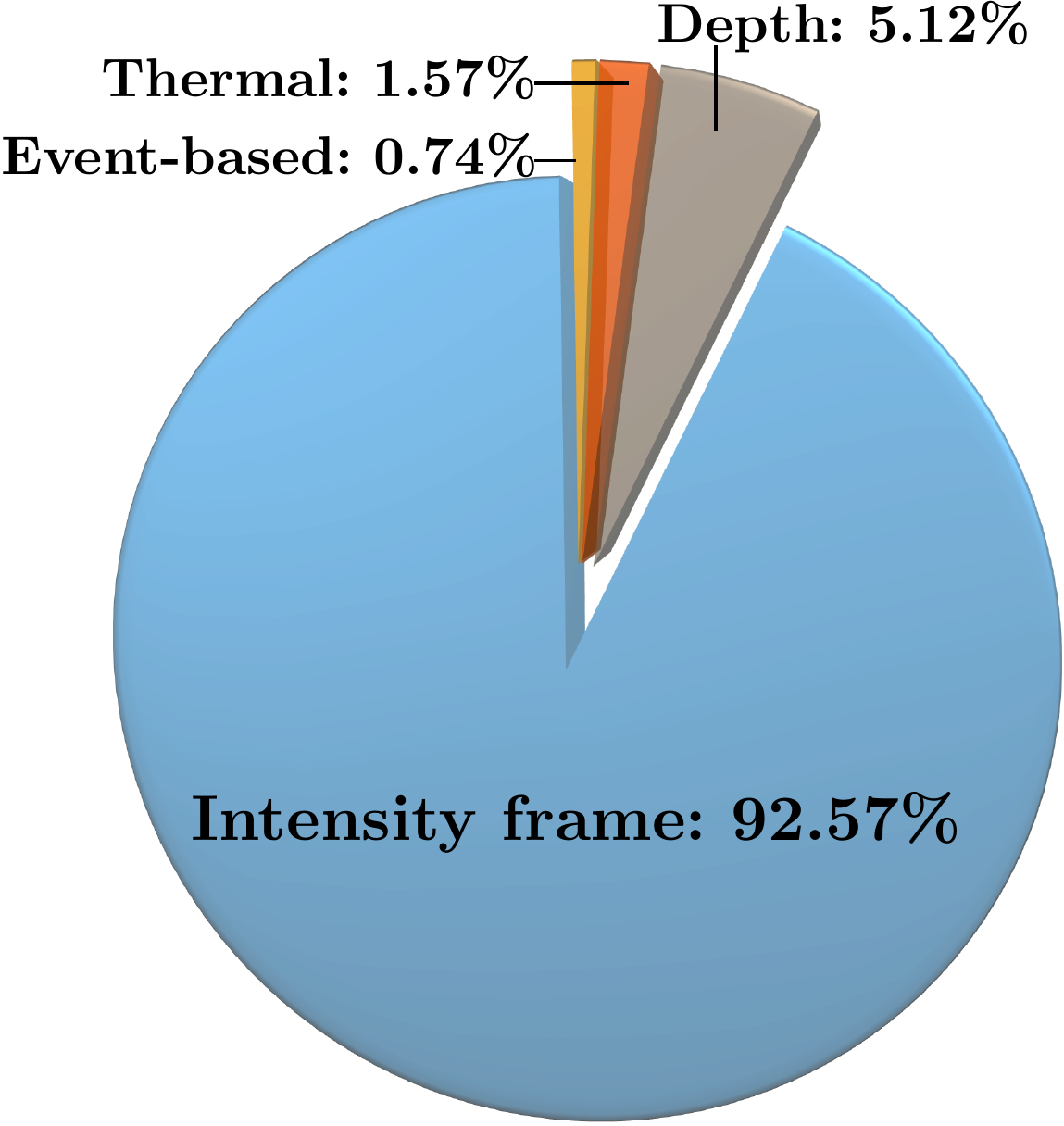}
  \end{center}
  \caption{Types of computer vision datasets.
  Data from \cite{cv:dataset:Fisher:2020}.}
  \label{fig:dataset:statistics}
\end{minipage}
\begin{minipage}[t]{0.55\linewidth}
    \begin{center}
    \includegraphics[width=\linewidth]{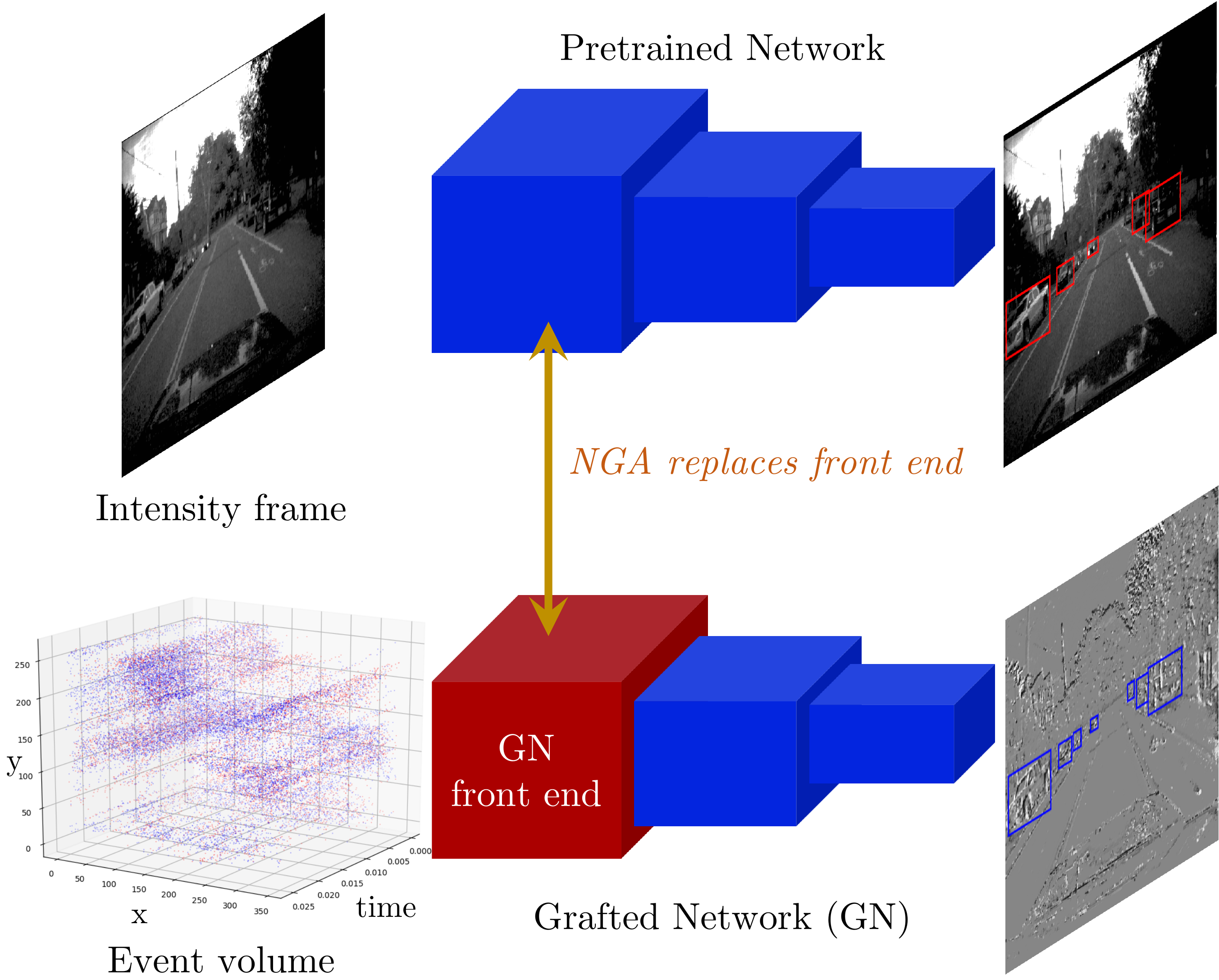}
    \end{center}
    \caption{A network 
    (blue) trained on intensity frames outputs bounding boxes of detected objects. 
    NGA trains a new GN front end (red) using a small 
    unlabeled dataset of recordings from a DAVIS~\cite{davis:Brandli:2014} event camera 
    that concurrently outputs intensity frames and asynchronous brightness change events. 
    The grafted network is obtained by \emph{replacing} the original front end with the GN front end, and is used for inference with the novel camera input data.
    }\label{fig:perception:transformation}    
\end{minipage}
\end{figure}

Novel vision sensors like thermal, hyperspectral, polarization, and event cameras provide new ways of sensing the visual world and enable new or improved vision system applications.
So-called \textit{event cameras}, for example, sense normal visible light, but dramatically sparsify it to pure brightness change events, 
which provide sub-ms timing and HDR to offer fast vision under challenging
illumination conditions~\cite{dvs:Lichtsteiner:2008,Gallego2019-EventCamSurvey}.
These novel sensors are becoming practical alternatives that complement standard cameras to improve vision systems.

Deep Learning (\textbf{DL}) with labeled data has revolutionized vision systems using conventional intensity frame-based cameras. 
But exploiting DL for vision systems based on novel cameras has been held back 
by the lack of large labeled datasets for these sensors. 
Prior work to solve high-level vision problems using inputs other than 
intensity frames has followed the principles of supervised Deep Neural Network (\textbf{DNN}) training algorithms,
where the task-specific datasets must be labeled with a tremendous amount 
of manual effort~\cite{predator:Moeys:2016,anumulatidigits18,thermal:segmentation:Bahnsen:2018,thermal:classification:Rodin:2018}. 
Although the community has collected many useful small datasets for novel sensors, 
the size, variety, and labeling quality of these datasets
is far from rivaling intensity frame datasets~\cite{nmnist:ncaltech:Orchard:2015,dvs:act:Hu:2016,anumulatidigits18,flir:2019,thermal:person:dataset:Kristo:2019,thermal:segmentation:Bahnsen:2018}. 
As shown in Fig.~\ref{fig:dataset:statistics},
among 1,212 surveyed computer vision datasets in~\cite{cv:dataset:Fisher:2020}, 
93\% are intensity frame 
datasets. Notably, 
there are only 28 event-based and thermal datasets. 

Particularly for event cameras, another line of DL research
employs unsupervised methods to train networks that
predict pixel-level quantities such as optical flow~\cite{event:voxel:Zhu:2019}, 
depth~\cite{depth:estimation:Zhu:2019};
and that reconstruct intensity frames~\cite{ev2vid:Rebecq:2019}. 
The information generated by these networks can be further 
processed by a downstream DNN trained to solve tasks such as object classification. This information is exceptionally useful in challenging scenarios
such as high-speed motion under difficult lighting conditions. The additional latency introduced by running these networks might be undesirable for fast online applications.
For instance, the DNNs used for intensity reconstruction at low QVGA resolution take $\sim$30\,ms on a dedicated GPU~\cite{ev2vid:Rebecq:2019,recon:firenet:Scheerlinck2020-wo}.

This paper introduces a simple yet effective algorithm called the
\emph{Network Grafting Algorithm} (\textbf{NGA}) to obtain a \emph{Grafted Network} (\textbf{GN}) that addresses both issues: 
1. the lack of large labeled datasets for training a DNN from scratch, and 
2. additional inference cost and latency that comes from running networks that  
compute pixel-level quantities.
With this algorithm, we train a
\emph{GN front end} for processing 
unconventional visual inputs (red block in Fig.~\ref{fig:perception:transformation}) 
to drive a network originally trained on intensity frames. We demonstrate
GNs for thermal and event cameras in this paper.

The NGA training encourages the GN front 
end to produce features that are similar 
to the features at several early layers of the pretrained network. 
Since the algorithm only requires pretrained hidden features
as the target, the training is self-supervised, that is, no labels are needed from the novel camera data.
The training method is described in Section~\ref{subsec:main:method}.
Furthermore, the newly trained GN has a similar inference cost 
to the pretrained network and does not introduce additional preprocessing latency.
Because the training of a GN front end relies on the pretrained network,
the NGA has similarities to 
Knowledge Distillation (KD)~\cite{knowledge:distill:Hinton:2015}, Transfer Learning~\cite{transfer:learning:Pan:2010}, and Domain Adaptation (DA)~\cite{da:Ganin:2015,da:Sun:2019,universal:da:You:2019}. In addition, our proposed algorithm
utilizes loss terms proposed for 
super-resolution image reconstruction and image style transfer~\cite{perceptual:loss:Johnson:2016,gram:loss:Gatys:2016}. 
Section~\ref{sec:related:works} 
elaborates on the similarities and differences between NGA and these related domains.

To evaluate NGA, we start with a pretrained object detection network 
and obtain a GN for a thermal object detection dataset (Section~\ref{subsec:thermal:results}) to solve the same task.
Then, we further demonstrate the training method 
on car detection using an event camera driving dataset (Section~\ref{subsec:mvsec:results}).
We show that the GN achieves similar 
detection precision compared to the original pretrained network. 
We also evaluate the accuracy gap between supervised and NGA self-supervised with MNIST for event cameras (Section~\ref{subsec:mnist}). Finally, we do representation analysis and ablation studies in Section~\ref{sec:analysis}.
Our contributions are as follows:
\begin{enumerate}
    \item We propose a novel algorithm called NGA that allows the use of
    networks already trained to solve a high-level vision problem  but adapted to work with a new GN 
    front end that processes inputs from thermal/event cameras.
    \item The NGA algorithm does not need a labeled thermal/event 
    dataset because the training is self-supervised.
    \item The newly trained GN has an inference 
    cost similar to the pretrained network because it directly processes 
    the thermal/event data. Hence, the computation latency 
    brought by \eg, intensity reconstruction from events is eliminated.
    \item The algorithm allows the output of these novel cameras to be exploited in situations that are difficult for standard cameras.
\end{enumerate}

\section{Related Work}\label{sec:related:works}

The NGA trains a GN front end such that the 
hidden features at different layers of the GN are similar 
to respective pretrained network features on intensity frames. 
From this aspect, the NGA is similar to 
Knowledge Distillation~\cite{knowledge:distill:Hinton:2015,fitnets:Romero:2015,knowledge:distill:Yim:2017}
where the knowledge of a teacher network is gradually distilled into a student network
(usually smaller than the teacher network) via the soft labels provided by the teacher network. 
In KD, the teacher and student networks use the same dataset. 
In contrast, the NGA assumes that the inputs
for the pretrained front end and the GN 
front end come from two \emph{different} modalities 
that see the same scene concurrently, 
but this dataset can simply be raw unlabeled recordings. The NGA is also a form of Transfer Learning~\cite{transfer:learning:Pan:2010}
and Domain Adaptation~\cite{da:Ganin:2015,da:Sun:2019,universal:da:You:2019} 
that study how to fine-tune the knowledge of a pretrained network on
a new dataset. 
\emph{Our method trains a GN front end 
from scratch since the network has to process the data from a different sensory modality.}

Another interpretation of maximizing hidden feature similarity
can be understood from the algorithms used for super-resolution (SR) 
image reconstruction and image style transfer. SR image reconstruction requires a network that up-samples a low-resolution 
image into a high-resolution image. The perceptual loss~\cite{perceptual:loss:Johnson:2016,perceptual:loss:Zhang:2018} 
was used to increase the sharpness and maintain the natural image 
statistics of the reconstruction. Image style transfer networks often 
aim to transfer an image into a target artistic style where Gram loss~\cite{gram:loss:Gatys:2016} is often employed.
While these networks learn to match either a high-resolution image ground 
truth or an artistic style, we train the GN front end to output features that match the hidden features of the pretrained network.
For training the front end, we draw inspiration from these 
studies and propose the use of combinations of training loss metrics including perceptual loss and Gram loss.

\section{Methods} \label{sec:methods}

We first describe the details of NGA in Section~\ref{subsec:main:method}, then the the event camera and its data representation in
Section~\ref{subsec:thermal:event:representation}.
Finally in Section~\ref{subsec:datasets}, we discuss the details of the thermal and event datasets.

\subsection{Network Grafting Algorithm} \label{subsec:main:method}

The NGA uses a pretrained network $\mathtt{N}$ that takes an intensity frame $I_{t}$ at time $t$,
and produces a grafted network $\mathtt{GN}$ whose input is a thermal frame or an event volume $V_{t}$.
$I_{t}$ and $V_{t}$ are synchronized during the training.
The $\mathtt{GN}$ should perform with similar accuracy on the same network task, such as object detection.
During inference with the thermal or event camera, $I_{t}$ is not needed.
The rest of this section sets up the constructions of $\mathtt{N}$ and $\mathtt{GN}$, then the NGA is described.

\begin{figure}[ht]
    \centering
    \includegraphics[width=0.95\linewidth]{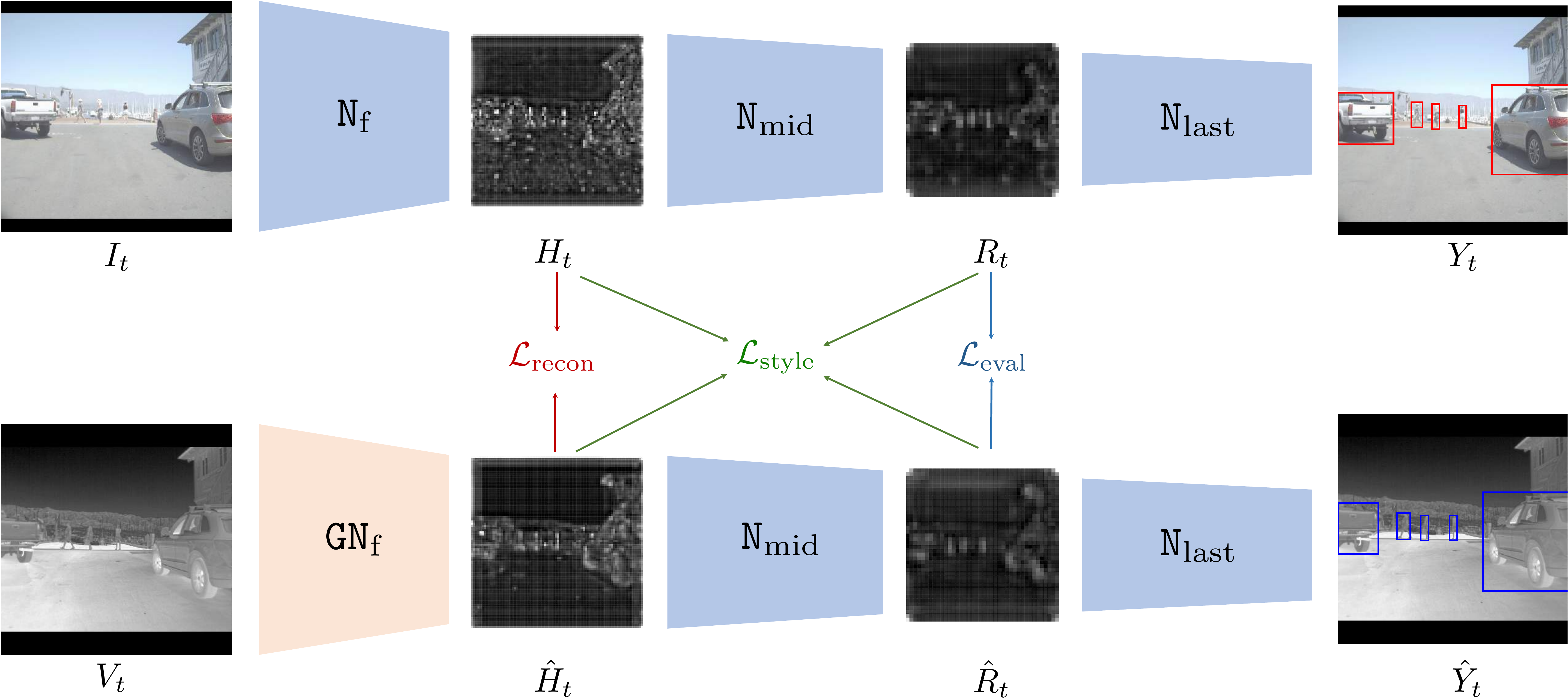}
    \caption{\textbf{NGA.} (\textbf{top}) Pretrained Network. (\textbf{bottom}) Grafted Network. Arrows point from variables to relevant loss terms. $I_{t}$ and $V_{t}$ here are an intensity frame and a thermal frame, respectively. The intermediate features $\hat{H}_{t}$, $H_{t}$, $\hat{R}_{t}$, $R_{t}$ are shown as heat maps averaged across channels. The object bounding boxes predicted by the original and the grafted network are outlined in red and blue correspondingly.
    }
    \label{fig:method:description}
\end{figure}

\noindent\textbf{Pretrained network setup.} The pretrained network $\mathtt{N}$ consists of three blocks: $\{\mathtt{N}_{\text{f}}$ (Front end), $\mathtt{N}_{\text{mid}}$ (Middle net), $\mathtt{N}_{\text{last}}$ (Remaining layers)$\}$.  Each block is made up of several layers and the outputs of each of the three blocks are defined as
\begin{equation}
    H_{t}=\mathtt{N}_{\text{f}}(I_{t}),\qquad R_{t}=\mathtt{N}_{\text{mid}}(H_{t}),\qquad Y_{t}=\mathtt{N}_{\text{last}}(R_{t})
\end{equation}
where $H_{t}$ is the front end features, $R_{t}$ is the middle net features, and $Y_{t}$ is the network prediction.
The separation of the network blocks is studied in Section~\ref{subsec:sepblocks}.
The top row in Fig.~\ref{fig:method:description} illustrates the three blocks of the pretrained network.

\noindent\textbf{Grafted network setup.} We define a \emph{GN front end} $\mathtt{GN}_{\text{f}}$ that takes $V_{t}$ as the input and outputs grafted front end features, $\hat{H}_{t}$,
of the same dimension as $H_{t}$.
$\mathtt{GN}_{\text{f}}$
combined with $\mathtt{N}_{\text{mid}}$ and $\mathtt{N}_{\text{last}}$ produces the
predictions $\hat{Y}$:
\begin{equation}
    \hat{H}_{t}=\mathtt{GN}_{\text{f}}(V_{t}),\qquad \hat{Y}_{t}=\mathtt{N}_{\text{last}}(\mathtt{N}_{\text{mid}}(\hat{H}_{t}))
\end{equation}
We define $\mathtt{GN}=\{\mathtt{GN}_{\text{f}}, \mathtt{N}_{\text{mid}}, \mathtt{N}_{\text{last}}\}$ as the \emph{Grafted Network} (bottom row of Fig.~\ref{fig:method:description}).

\noindent\textbf{Network Grafting Algorithm.} The NGA trains the grafted
network $\mathtt{GN}$ to reach a similar performance to that of the
pretrained network $\mathtt{N}$ by increasing the representation similarity between features $H=\{H_{t}|\forall t\}$ and $\hat{H}=\{\hat{H}_{t}|\forall t\}$.

The loss function for the training of the $\mathtt{GN}_{\text{f}}$ consists of a combination of three losses.
The first loss is the Mean-Squared-Error (MSE) between $H$ and $\hat{H}$:
\begin{equation}
    \mathcal{L}_{\text{recon}}=\text{MSE}(H, \hat{H})
\end{equation}
Because this loss term captures the amount of representation similarity between the two different
front ends, we call $\mathcal{L}_{\text{recon}}$ a \emph{Feature Reconstruction Loss} (FRL).

The second loss takes into account the output of the middle net layers in the network and draws inspiration from the Perception Loss~\cite{perceptual:loss:Johnson:2016}.
This loss is set by the MSE between the middle net frame features $R=\{R_{t}|\forall t\}$ and the grafted middle net features $\hat{R}=\{\mathtt{N}_{\text{mid}}(\hat{H}_{t})|\forall t\}$:
\begin{equation}
    \mathcal{L}_{\text{eval}}=\text{MSE}(R, \hat{R})
\end{equation}
Since this loss term additionally evaluates the feature similarities between front end features $\{H, \hat{H}\}$, we refer to $\mathcal{L}_{\text{eval}}$ as the \emph{Feature Evaluation Loss} (FEL).

Both FRL and FEL terms minimize the magnitude differences between hidden features. To further encourage the GN front end to generate intensity frame-like textures, we introduce the \emph{Feature Style Loss} (FSL) based on the mean-subtracted Gram loss~\cite{gram:loss:Gatys:2016} that computes a Gram matrix using feature columns across channels (indexed using $i$, $j$). The Gram matrix represents image texture rather than spatial structure. This loss is defined as:
\begin{align}
    &\text{Gram}(F)^{(i,j)}=\sum_{\forall t}\tilde{F}_{t}^{(i)\top}\tilde{F}_{t}^{(j)},\quad\text{where }  \tilde{F}_{t}=F_{t}-\text{mean}(F_{t}) \\
    &\mathcal{L}_{\text{style}}=\gamma_{h}\text{MSE}(\text{Gram}(H), \text{Gram}(\hat{H}))+\gamma_{r}\text{MSE}(\text{Gram}(R), \text{Gram}(\hat{R}))
\end{align}
The final loss function is a weighted sum of the three loss terms:
\begin{equation}
    \mathcal{L}_{\text{tot}}=\alpha\mathcal{L}_{\text{recon}}+\beta\mathcal{L}_{\text{eval}}+\mathcal{L}_{\text{style}}
\end{equation}
For all experiments in the paper, we set $\alpha=\beta=1$, $\gamma_{h}\in\{10^{5}, 10^{6}, 10^{7}\}$, $\gamma_{r}=10^{7}$. The loss terms and their associated variables are
shown in Fig.~\ref{fig:method:description}. The importance of each loss term is studied in Section~\ref{subsec:loss}.

\subsection{Event Camera and Feature Volume Representation} \label{subsec:thermal:event:representation}

Event cameras such as the DAVIS camera~\cite{dvs:Lichtsteiner:2008,davis:Brandli:2014} produce a stream of asynchronous ``events'' triggered by local brightness (log intensity) changes at individual pixels.
Each output event of the event camera is a four-element tuple $\{t, x, y, p\}$
where $t$ is the timestamp, $(x, y)$ is the location of the event,
and $p$ is the event polarity. The polarity is either positive (brightness increasing) or negative (brightness decreasing). To preserve both spatial and temporal information captured by the polarity events, we use 
the event voxel grid~\cite{event:voxel:Zhu:2019,ev2vid:Rebecq:2019}.
Assuming a volume of $N$ events $\{(t_{i}, x_{i}, y_{i}, p_{i})\}_{i=1}^{N}$ where $i$ is the event index, we divide this volume into $D$ event slices of equal temporal intervals such that the $d$-th slice $S_{d}$ is defined as follows:
\begin{equation}
    \forall x, y;\quad S_{d}(x, y)=\sum_{x_{i}=x, y_{i}=y}p_{i}\max(0, 1-|d-\tilde{t}_{i}|)
\end{equation}
and $\tilde{t}_{i}=(D-1)\frac{t_{i}-t_{1}}{t_{N}-t_{1}}$ is the normalized event timestamp. The event volume is then defined as $V_{t}=\{S_{d}\}_{d=0}^{D-1}$. In Section~\ref{sec:experiments}, $D=3,10$ and $N=25,000$.
Prior work has shown that this spatio-temporal view of the input scene activity
covering a constant number of brightness change events 
is simple but effective for optical flow computation~\cite{event:voxel:Zhu:2019} and video reconstruction~\cite{ev2vid:Rebecq:2019}.

\subsection{Datasets} \label{subsec:datasets}

Two different vision datasets were used in the experiments in this paper and are presented in the subsections.

\noindent\textbf{Thermal dataset for object detection.} The FLIR Thermal Dataset~\cite{flir:2019} includes labeled recordings from a thermal camera for driving on Santa Barbara, CA area streets and highways for both day and night.
The thermal frames were captured using a FLIR IR Tau2 thermal camera with an image resolution of 640$\times$512.
The dataset has parallel RGB intensity frames and thermal frames in an 8-bit JPEG format with AGC. Since the standard camera is placed alongside the thermal camera, a constant spatial displacement is expected, and this shift is corrected for the training samples. The dataset has 4,855 training intensity-thermal pairs, and 1,256 testing pairs, of which 60\% are daytime and 40\% are nighttime driving samples.
We excluded samples where the intensity frames are corrupted. The annotated object classes are \texttt{car}, \texttt{person}, and \texttt{bicycle}.

\noindent\textbf{Event camera dataset.}
The Multi Vehicle Stereo Event Camera Dataset (MVSEC)~\cite{mvsec:Zhu:2018} is a collection of event camera recordings for studying 3D perception and optical flow estimation. The \texttt{outdoor\_day2} recording is carried out in an urban area of West Philadelphia. This recording was selected for the car detection experiment because of its better quality compared to other recordings, and it has a large number of cars in the scenes distributed throughout the entire recording.
We generated in total 7,000 intensity frames and event volume pairs from this recording. Each event volume contains $N=25,000$ events.
The first 5,000 pairs are used as the training dataset, and the last 2,000 pairs are used as the testing dataset. There are no temporally overlapping pairs between the training and testing datasets. 

Because MVSEC does not provide ground truth bounding boxes for cars, we pseudo-labeled data pairs of the testing dataset for intensity frames that contain at least one car detected by the Hybrid Task Cascade (HTC) Network~\cite{htc:network:Chen:2019}, which provides state-of-the-art results in object detection. We only use the bounding boxes with 80\% or higher confidence to obtain high-quality bounding boxes.
To compare the effect of using different numbers of event slices $D$ in an event volume on the detection results, we additionally created two versions of this dataset: DVS-3 where $D=3$ and DVS-10 where $D=10$.

\section{Experiments} \label{sec:experiments}

We use the NGA to train a GN front end for a pretrained object detection network. In this case, we use the YOLOv3 network~\cite{yolov3:Redmon:2018} that was trained using the COCO dataset~\cite{coco:dataset:Lin:2014} with 80 objects.
This network was chosen because it still provides good detection 
accuracy and could be deployed on a low-cost embedded real-time platform.
The pretrained network is referred to as YOLOv3-$\mathtt{N}$ and the grafted thermal/event-driven networks as YOLOv3-$\mathtt{GN}$ in the rest of the paper. The training inputs consist of $224\times224$ image patches randomly cropped from the training pairs. No other data augmentation is performed. All networks are trained for 100 epochs with the Adam optimizer~\cite{adam:Kingma:2014}, a learning rate of $10^{-4}$, and a mini-batch size of 8. Each model training takes $\sim$1.5 hours using an NVIDIA RTX 2080 Ti, which is only 5\% of the 2 days it typically requires to train one of the object detectors used in this paper on standard labeled datasets.
More results from the experiments on the different vision datasets are presented in the supplementary material.

\subsection{Object Detection on Thermal Driving Dataset}\label{subsec:thermal:results}

This section presents the experimental results of using the NGA to train an object detector for the thermal driving dataset.

\begin{figure*}[ht]
    \centering
    \includegraphics[width=\linewidth]{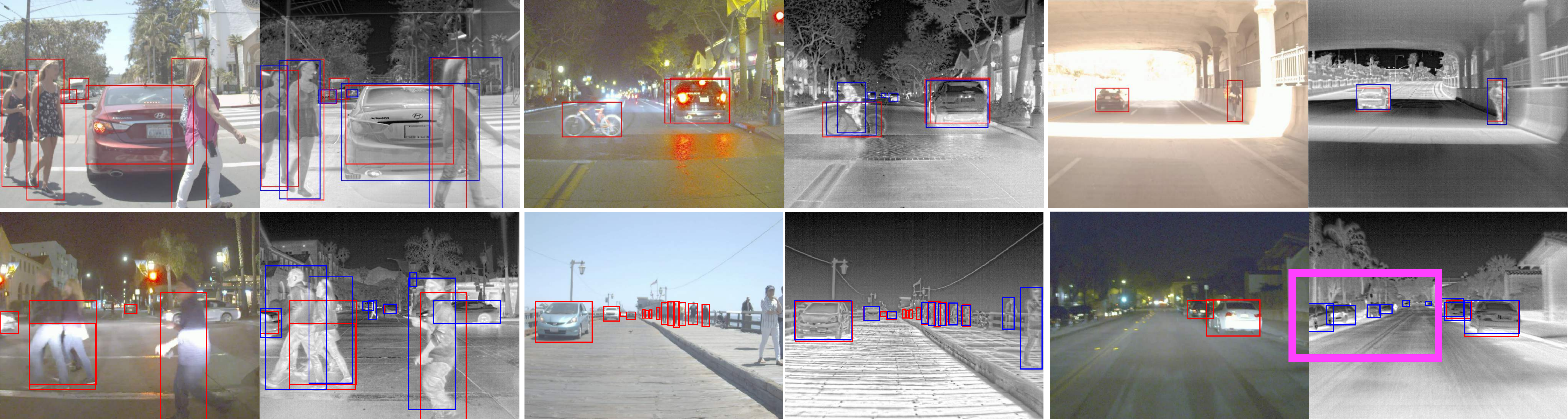}
   \caption{Examples of six testing pairs from the thermal driving dataset. The red boxes are objects detected by the original intensity-driven YOLOv3 network and the blue boxes show the objects detected by the thermal-driven network. The magenta box shows cars detected by the thermal-driven GN that are missed by the intensity-driven network when the intensity frame is underexposed. Best viewed in color.}
\label{fig:training:examples:thermal}
\end{figure*}
 
Fig.~\ref{fig:training:examples:thermal} shows six examples of object detection results from the original intensity-driven YOLOv3 network and the thermal-driven network.
These examples show that when the intensity frame is well-exposed, the prediction difference between YOLOv3-\texttt{N} and YOLOv3-\texttt{GN} appears to be small.
However, when the intensity frame is either underexposed or noisy, the thermal-driven network detects many more objects than the pretrained network.
For instance, in the magenta box of Fig.~\ref{fig:training:examples:thermal}, most cars are not detected by the intensity-driven  network but they are detected by the thermal-driven network.

The detection precision (AP$_{50}$) results over the entire test set (Table~\ref{tab:obj:detect:result:flir}) show that the accuracy of our pretrained YOLOv3-$\mathtt{N}$ on the intensity frames (30.36)
is worse than on thermal frames (39.92) because 40\% of the intensity night frames look noisy and are underexposed.
The YOLOv3-$\mathtt{GN}$ thermal-driven network achieved the highest AP$_{50}$ detection precision (45.27) among all our YOLOv3 variants while requiring training of only 5.17\% (3.2M) parameters with NGA. A baseline Faster R-CNN which was trained on the same labeled thermal dataset~\cite{flir:2019} achieved a higher precision of 53.97. However, it required training of 47M parameters  which is 15X more than the YOLOv3-$\mathtt{GN}$.
Overall, the results show that the self-supervised GN front end significantly improves the accuracy results of the original network on the thermal dataset. 

\begin{table}[ht]
    \centering
    \caption{Object detection AP$_{50}$ scores on the FLIR driving dataset. The training of YOLOv3-$\mathtt{GN}$ repeats five times.}
    \label{tab:obj:detect:result:flir}
    \begin{tabular}{l|l|c|c}
        \hline
        \multicolumn{1}{c|}{\textbf{Network}} & \multicolumn{1}{c|}{\textbf{Modality}} &  AP$_{50}$  & \#~Trained Params \\
        \hline
        \multicolumn{4}{c}{This work} \\
        \hline
        YOLOv3-$\mathtt{GN}$ & Thermal & \textbf{45.27$\pm$1.14} & \textbf{3.2M}\\
        YOLOv3-$\mathtt{N}$ & Intensity & 30.36 & 62M \\
        YOLOv3-$\mathtt{N}$ & Thermal & 39.92 & 62M\\
        \hline
        SSD & Intensity & 8.00 & 36M\\
        Faster R-CNN & Intensity & 23.82 & 42M\\
        Cascade R-CNN & Intensity & 27.10  & 127M\\
       \hline
        \multicolumn{4}{c}{Baseline supervised thermal object detector} \\
        \hline
        Faster R-CNN~\cite{flir:supervised:Devaguptapu:2019} & Thermal & 53.97 & 47M\\
         \hline
    \end{tabular}
\end{table}

For comparison with other object detectors, we also use the \texttt{mmdetection} framework~\cite{mmdetection:2019} to process the intensity frames using pretrained SSD~\cite{ssd:Liu:2016}, Faster R-CNN~\cite{faster:rcnn:Ren:2015} and Cascade R-CNN~\cite{cascade:rcnn:Cai:2018} detectors. All have worse AP$_{50}$ scores than any of the YOLOv3 networks, so 
YOLOv3 was a good choice for evaluating the effectiveness of NGA.

\subsection{Car Detection on Event Camera Driving Dataset}\label{subsec:mvsec:results}

To study if the NGA is also effective for exploiting another visual sensor, \eg, an event camera, 
we evaluated car detection results using the pretrained network YOLOv3-$\mathtt{N}$ and a grafted network YOLOv3-$\mathtt{GN}$ using the MVSEC dataset.

\begin{figure}[ht]
    \centering
    \includegraphics[width=\linewidth]{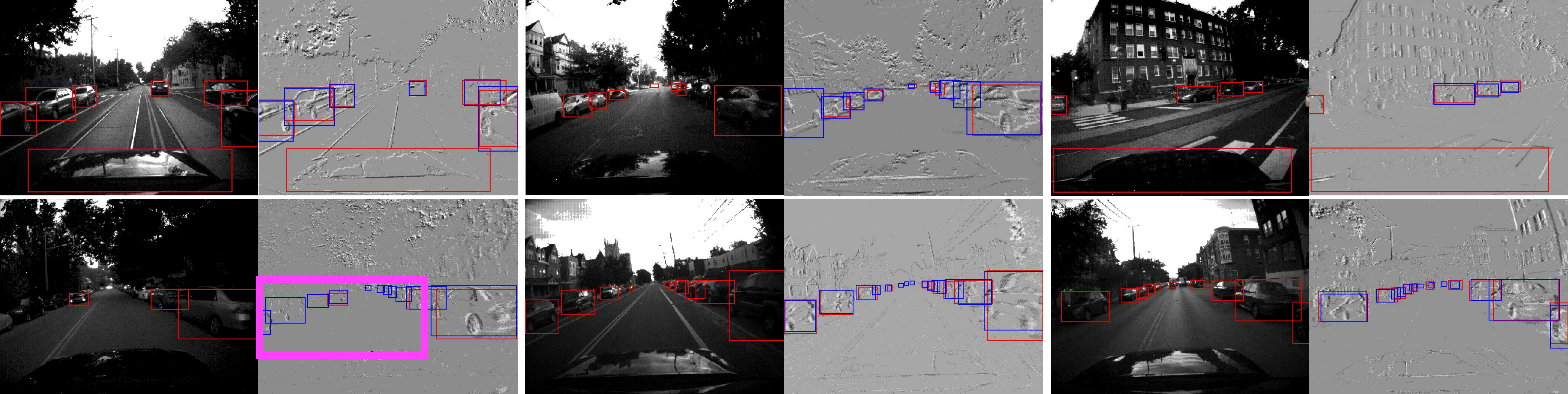}
   \caption{Examples of testing pairs from the MVSEC dataset. The event volume is visualized after averaging across slices. The predicted bounding boxes (in red) from the intensity-driven network can be compared with the predicted bounding boxes (in blue) from the event-driven network. The magenta box shows cars detected by the event-driven network that are missed by the intensity-driven network. Best viewed in color.}
\label{fig:training:examples}
\end{figure}

The event camera operates over a larger dynamic range of lighting than an intensity frame camera and therefore will detect moving objects even in poorly lighted scenes. From the six different data pairs in the MVSEC testing dataset (Fig.~\ref{fig:training:examples}), we see that the event-driven YOLOv3-$\mathtt{GN}$ network detects most of the cars found in the intensity frames and additional cars not detected in the intensity frame (see the magenta box in the figure).
These examples help illustrate how event cameras and the event-driven network can complement the pretrained network in challenging situations. 

Table~\ref{tab:obj:detect:result:mvsec} compares the accuracy of the intensity and event camera detection networks on the testing set. As might be expected for these well-exposed and sharp daytime intensity frames, the YOLOv3-$\mathtt{N}$ produces the highest average precision (AP).
Surprisingly, the YOLOv3-$\mathtt{GN}$ with DVS-3 input achieves close to the same accuracy, although it was never explicitly trained to detect objects on this type of data.
We also tested if the pretrained network would perform poorly on the DVS-3 event dataset.
The AP$_{50}$ is almost 0
(not reported in the table) and confirms that the intensity-driven front end fails at processing the event volume and that using a GN front end is essential for acceptable accuracy.

We also compare the performances of the event-driven networks that receive as input, the two datasets with different numbers of event slices for the event volume, \emph{i.e.}, DVS-3, and DVS-10. The network trained on DVS-10 shows a better score of AP$_{50}=70.35$, which is only 3.18 lower than the original YOLOv3 network accuracy.
Table~\ref{tab:obj:detect:result:mvsec} also shows 
the effect on accuracy when varying the number of training samples. Even when trained using only 40\% of training data (2k samples), the YOLOv3-$\mathtt{GN}$ still shows strong detection precision at 66.75. But when the NGA has access to only 10\% of the data (500 samples) during training, the detection performance drops by 22.47\% compared to the best event-driven network. Although the NGA requires far less data compared to standard supervised training, training with only a few hundreds of samples remains challenging and could benefit from data augmentation to improve performance.

\begin{table}[ht]
    \centering
    \caption{AP$_{50}$ scores for car detection on the MVSEC driving dataset (five runs).}
    \label{tab:obj:detect:result:mvsec}
    \begin{tabular}{l|l|c|c}
        \hline
        \multicolumn{1}{c|}{\textbf{Network}} & \multicolumn{1}{c|}{\textbf{Modality}} &  AP$_{50}$ & \#~Trained Params\\
        \hline
        YOLOv3-$\mathtt{N}$ & Intensity & 73.53 & 62M\\
        \hline
        YOLOv3-$\mathtt{GN}$ & DVS-3 & 70.14$\pm$0.36 & 3.2M\\
        YOLOv3-$\mathtt{GN}$ & DVS-10 & 70.35$\pm$0.51 & 3.2M\\
        YOLOv3-$\mathtt{GN}$ & DVS-10 (40\% samples) & 66.75$\pm$0.30 & 3.2M\\
        YOLOv3-$\mathtt{GN}$ & DVS-10 (10\% samples) & 47.88$\pm$1.86 & 3.2M\\
        \hline
        \textbf{Combined} & Intensity+DVS-10 & \textbf{75.45} & N/A\\
        \hline
        SSD & Intensity & 36.17 & 36M\\
        Faster R-CNN & Intensity & 71.89 & 42M \\
        Cascade R-CNN & Intensity & 85.16 & 127M\\
        \hline
    \end{tabular}
\end{table}

To study the benefit of using the event camera brightness change events to complement its intensity frame output, 
we combined the detection results from both the pretrained network and event-driven network (Row \textbf{Combined} in Table~\ref{tab:obj:detect:result:mvsec}).
After removing duplicated bounding boxes through non-maximum suppression, the AP$_{50}$ score of the combined prediction is higher by 1.92 than the prediction of the pretrained network using intensity frames.

Reference AP$_{50}$ scores from three additional intensity frame detectors implemented using the \texttt{mmdetection} toolbox are also reported in the table for comparison.

\subsection{Comparing NGA and Standard Supervised Learning}
\label{subsec:mnist}

Intuitively, a network trained in a supervised manner should perform better than a network trained through self-supervision.
To study this, we evaluate the accuracy gap between classification networks trained with supervised learning, and the NGA using event recordings of the MNIST handwritten digit recognition dataset, also called N-MNIST dataset~\cite{nmnist:ncaltech:Orchard:2015}.
Each event volume is prepared by setting $D=3$. The training uses the Adam optimizer, a learning rate of $10^{-3}$ and a batch size of 256.

First, we train the LeNet-\texttt{N} network with the standard LeNet-5 architecture~\cite{lenet:LeCun:1998} using the intensity samples in the MNIST dataset. Next, we train LeNet-\texttt{GN} with the NGA by using parallel MNIST and N-MNIST sample pairs.
We also train an event-driven LeNet-\texttt{supervised} network from scratch on N-MNIST using standard supervised learning with the labeled digits.
The results in Table~\ref{tab:classification} show that the accuracy of the LeNet-\texttt{GN} network is only 0.36\% lower than that of the event-driven LeNet-\texttt{supervised} network even with the training of a front end which has only 8\% of the total network parameters, and without the availability of labeled training data.
The LeNet-\texttt{GN} also performed better or on par with other models that have been tested on the N-MNIST dataset~\cite{hots:nmnist:Lagorce:2017,hfirst:nmnist:Ochard:2015,hats:nmnist:Sironi:2018}.

\begin{table}[ht]
    \centering
    \caption{Classification results on MNIST and N-MNIST datasets.}
    \label{tab:classification}
     \begin{tabular}{l|l|c|c}
        \hline
        \multicolumn{1}{c|}{\textbf{Network}} & \multicolumn{1}{c|}{\textbf{Dataset}} &  Error Rate (\%) & \#~Trained Params\\
        \hline
        LeNet-\texttt{N} & MNIST & 0.92 & 64k\\
        LeNet-\texttt{GN} & N-MNIST & 1.47$\pm$0.05 & 5k\\
        LeNet-\texttt{supervised} & N-MNIST & 1.11$\pm$0.06 & 64k\\
        \hline
        HFirst~\cite{hfirst:nmnist:Ochard:2015} & N-MNIST & 28.80 & - \\
        HOTS~\cite{hots:nmnist:Lagorce:2017} & N-MNIST & 19.20 & -\\
        HATS~\cite{hats:nmnist:Sironi:2018} & N-MNIST & 0.90 & - \\
        \hline
    \end{tabular}
\end{table}

\section{Network Analysis} \label{sec:analysis}

To understand the representational power of the GN features,
Section~\ref{subsec:decode:hidden} presents a qualitative study that shows how the grafted front end features represent useful visual input under difficult lighting conditions. To design an effective GN, it is important to select what parts of the network to graft. Sections~\ref{subsec:sepblocks} and \ref{subsec:loss} describe studies on the network variants and the importance of the loss terms.

\subsection{Decoding Grafted Front End Features}\label{subsec:decode:hidden}

Previous experiments show that the grafted front end features provide useful information for the GN in the object detection tasks. In this section, we provide qualitative evidence that the grafted features often faithfully represent the input scene. Specifically, we decode the grafted features by optimizing a decoded intensity frame $I_{t}^{\text{d}}$ that produces features through the intensity-driven network best matching the grafted features $\hat{H}_{t}$, by minimizing:
\begin{equation}
\text{arg}\min_{I_{t}^{\text{d}}}\text{MSE}(\mathtt{N}_{\text{f}}(I_{t}^{\text{d}}), \hat{H}_{t})+5\times\text{TV}(I_{t}^{\text{d}})
\end{equation}
where $\text{TV}(\cdot)$ is a total variation regularizer for encouraging spatial smoothness~\cite{tv:regularizer:Aly:2005}. The decoded intensity frame $I_{t}^{\text{d}}$ is initialized randomly and has the same spatial dimension as the intensity frame, then the pixel values of $I_{t}^{\text{d}}$ are optimized for 1k iterations using an Adam optimizer with learning rate of $10^{-2}$.

\begin{figure}[ht]
    \centering
    \includegraphics[width=\linewidth]{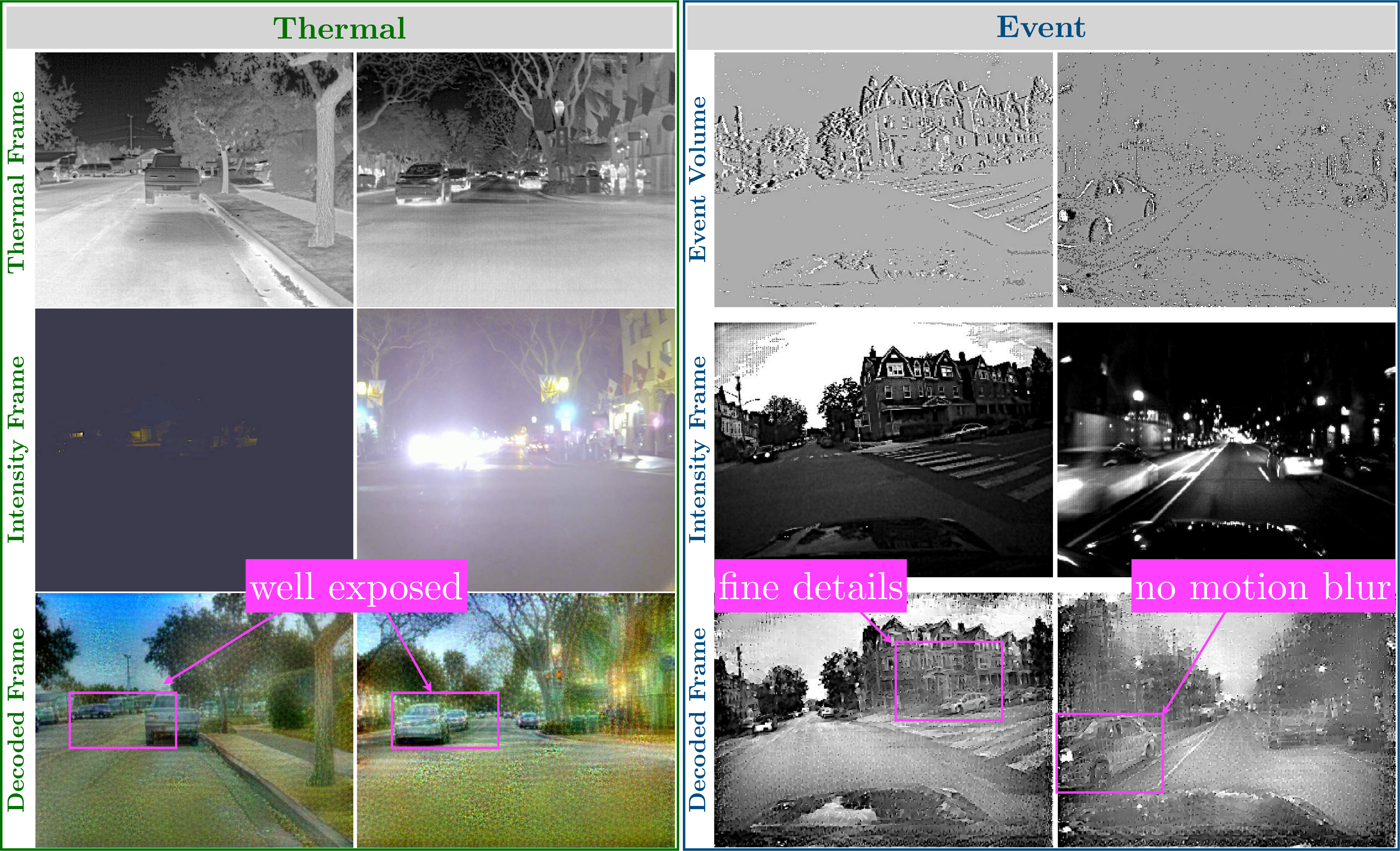}
    \caption{Decoded frames of image pairs taken from both the thermal and event datasets. Each column represents an example image from either the thermal dataset (the leftmost two columns) or the event dataset (the rightmost two columns). The top panel of each column shows either the thermal frame or the event volume.
    The middle panel shows the raw intensity frames. The bottom panel shows the decoded intensity frames (see main text).
    Labeled regions in the decoded frames show details that are not visible in the four original intensity frames. The figure is best viewed in color.}
    \label{fig:decode:representation}
\end{figure}

Figure~\ref{fig:decode:representation} shows four examples from the thermal dataset and the event dataset. Under extreme lighting conditions, the intensity frames are often under/over-exposed while the decoded intensity frames show that the thermal/event front end features can represent the same scene better (see the labeled regions).

\subsection{Design of Grafted Network}
\label{subsec:sepblocks}
\begin{wrapfigure}{r}{0.41\linewidth}
    \centering
    \includegraphics[width=\linewidth]{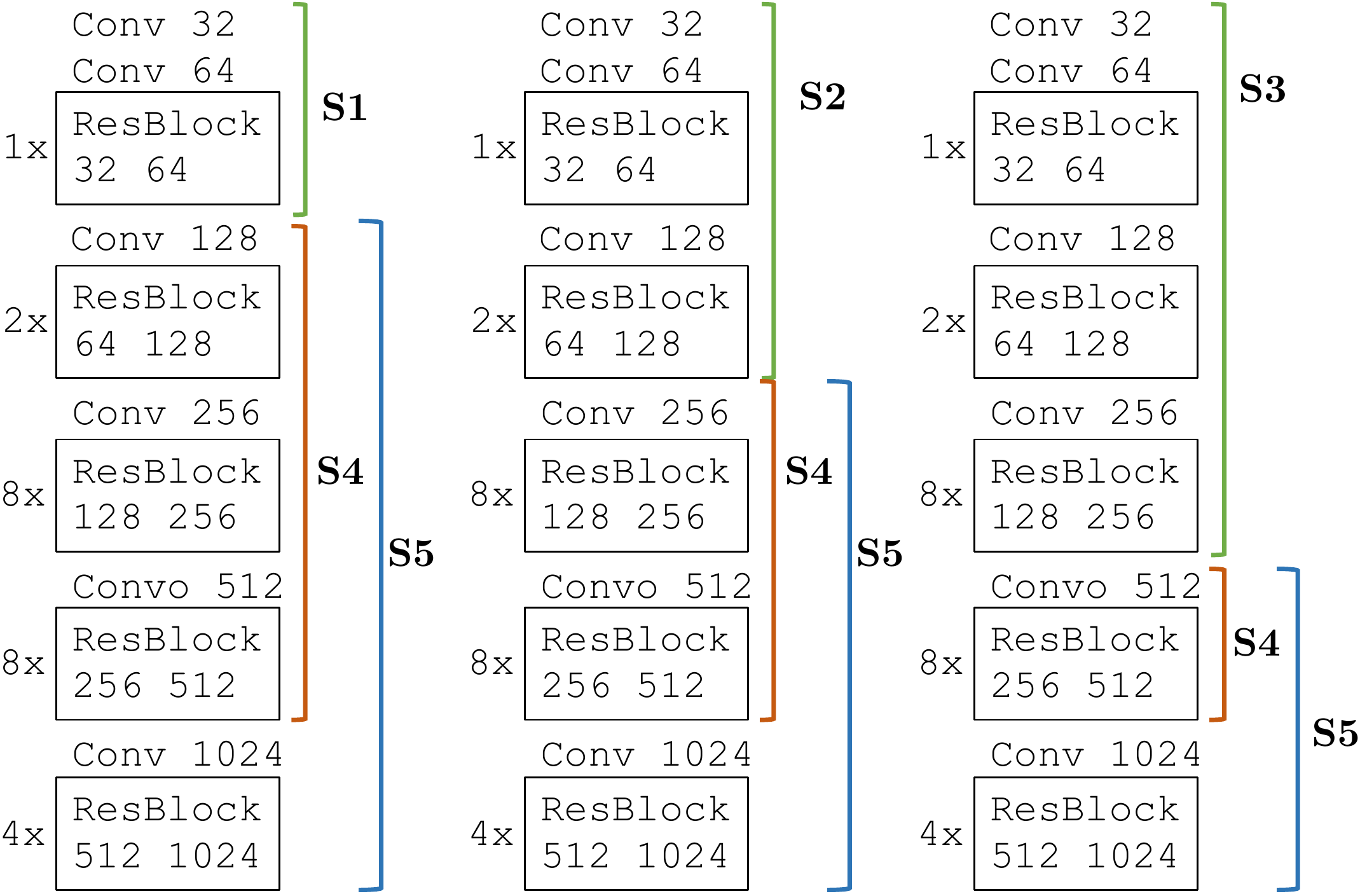}
    \caption{YOLOv3 backbone: Darknet-53~\cite{yolov3:Redmon:2018}. The front end variants are S1, S2 and S3. The middle net variants are S4 and S5. \texttt{Conv} represents a convolution layer, \texttt{ResBlock} represents a residual block.}
    \label{fig:darknet}
\end{wrapfigure}

\begin{figure}[ht]
    \centering
    \includegraphics[width=\linewidth]{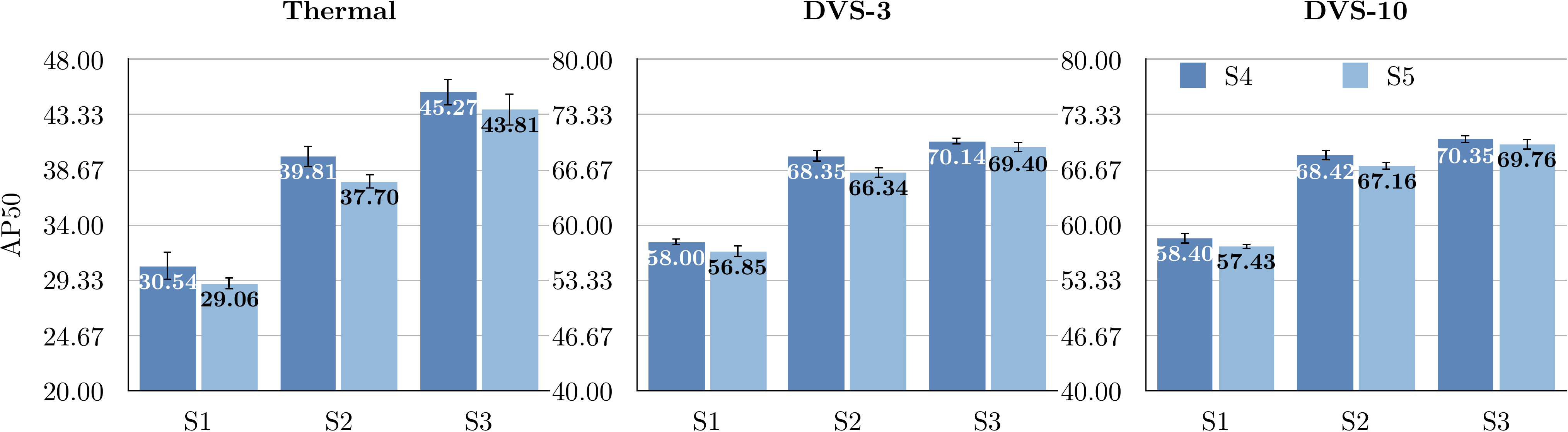}
    \caption{Results of different front end and middle net variants in Fig.~\ref{fig:darknet} for both thermal and event datasets in AP$_{50}$. Experiments for each variant are repeated five times.}
    \label{fig:blocks:separation}
\end{figure}

The backbone network of YOLOv3 is called Darknet-53, and consists of five residual blocks
(Fig.~\ref{fig:darknet}).
Selecting the correct set of residual blocks used for the NGA front end is important.
Six combinations of the front end and middle net by using different numbers of residual blocks: \{S1, S4\}, \{S1, S5\}, \{S2, S4\}, \{S2, S5\},
\{S3, S4\} and \{S3, S5\} are tested.
S1, S2, S3 indicate front end variants 
with different number of residual blocks that uses 0.06\% (40k), 0.45\% (279k), and 5.17\% (3.2M) of total parameters (62M) respectively. The number of blocks for S4 and S5 vary depending on the chosen variant.
Figure~\ref{fig:blocks:separation} shows the AP$_{50}$ scores for different combinations of front end and middle net variants. The best separation of the network blocks is \{S3, S4\}.
In the YOLOv3 network, the detection results improve sharply when the front end includes more layers. On the other hand, the difference in AP$_{50}$ between using S4 or S5 for the middle net is not significant.
These results suggest that using a deeper front end is better than a shallow front end, especially when training resources are not a constraint.

\subsection{Ablation Study on Loss Terms} \label{subsec:loss}

\begin{figure}[ht]
    \centering
    \includegraphics[width=\textwidth]{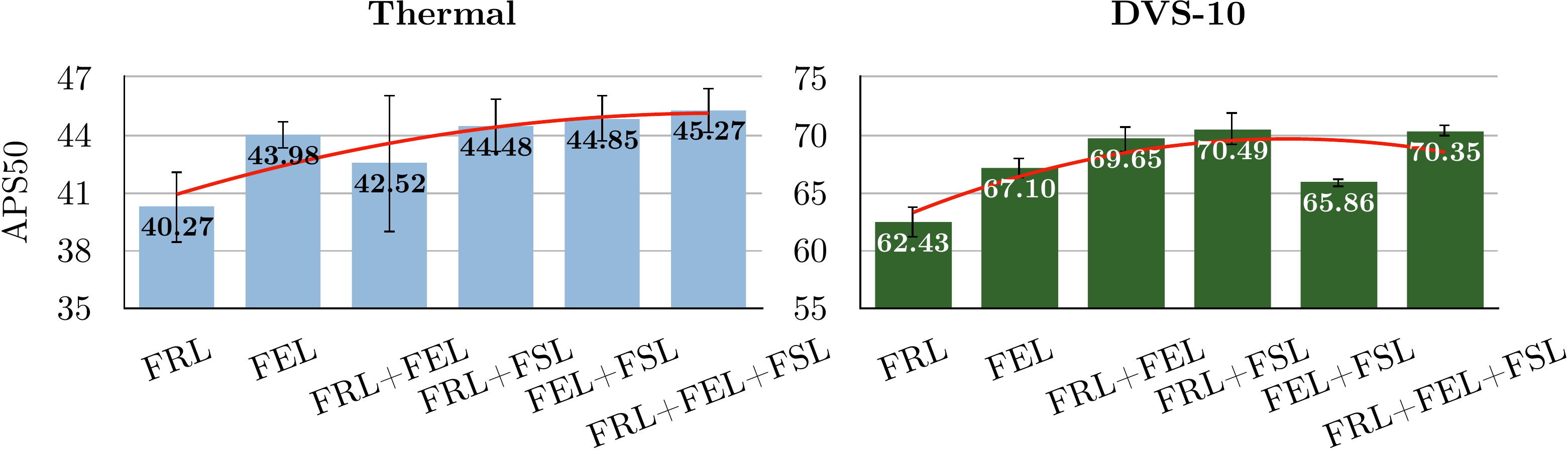}
    \caption{GN performance (AP$_{50}$) trained with different loss configurations. Results are from five repeats of each loss configuration.}
    \label{fig:ablation}
\end{figure}

The NGA training includes three loss terms: FRL, FEL, and FSL. We studied the importance of these loss terms by performing an ablation study using both the thermal dataset and the event dataset. These experiments are done on the network configuration \{S3, S4\} that gave the best accuracy (see Fig.~\ref{fig:blocks:separation}).
The detection precision scores are shown in Fig.~\ref{fig:ablation} for different loss configurations.
The FRL and the FEL are the most critical loss terms, while the role of the FSL is less significant.
The effectiveness of different loss combinations seems task-dependent and sometimes fluctuates, \eg, FRL+FEL for thermal and FEL+FSL for DVS-10.
The trend lines indicate that using a combination of loss terms is most likely to produce better detection scores.

\section{Conclusion} \label{sec:conclusion}
This paper proposes the Network Grafting Algorithm (NGA) that replaces the front end of a network that is pretrained on a large labelled dataset so that the new grafted network (GN) also works well with a different sensor modality.
Training the GN front end for a different modality, in this case, a thermal camera or an event camera, requires only a reasonably small unlabeled dataset ($\sim$\,5k samples) that has spatio-temporally synchronized data from both modalities. By comparison, the COCO dataset on which many object detection networks are trained has 330k images.  
Ordinarily, training a network with a new sensor type and limited labeled data requires a lot of careful data augmentation.
NGA avoids this by exploiting the new sensor data even if unlabeled because the pretrained network already has informative features.

The NGA was applied on an object detection network that was pretrained on a big image dataset. The NGA training was conducted using the FLIR thermal dataset~\cite{flir:2019} and the MVSEC driving dataset~\cite{mvsec:Zhu:2018}.
After training, the GN reached a similar or higher average precision (AP$_{50}$) score compared to the precision achieved by the original network.
Furthermore, the inference cost of the GN is similar to that of the pretrained network, which eliminates the latency cost for computing low-level quantities, particularly for event cameras. 
This newly proposed NGA widens the use of these unconventional cameras to a broader range of computer vision applications.

\noindent\textbf{Acknowledgements.}
This work was funded by the Swiss National Competence Center in Robotics (NCCR Robotics).

%
%
\bibliographystyle{splncs04}
\bibliography{egbib}
\end{document}